# Data Quality Evaluation using Probability Models

Allen O'Neill[1]

[1]Department of Computer Science, Reading University, Reading, RG6 6AY, United Kingdom
allen.oneill@pgr.reading.ac.uk

Abstract: This paper discusses an approach with machine-learning probability models to evaluate the difference between good and bad data quality in a dataset. A decision tree algorithm is used to predict data quality based on no domain knowledge of the datasets under examination. It is shown that for the data examined, the ability to predict the quality of data based on simple good/bad pre-labelled learning examples is accurate, however in general it may not be sufficient for useful production data quality assessment.

## 1 RESEARCH PROBLEM

With any form of decision making, the quality of the decision is only as good as the data analysed, and so there is a need to ensure that the highest quality data is available for analysis. While having good levels of data quality improves analysis accuracy, having bad data quality can have a serious impact on the enterprise. A UK industry report (Vision, 2009) estimated that bad quality data costs the UK retail industry over £700k per annum. This report also provided a cause-and-effect analysis to illustrate where data quality could be improved, and the effect this would have on resulting commercial outcomes. A further report (Schutz, 2016) indicates that 58% of organisations surveyed had an unsophisticated approach to data quality, being unaware of the impact of data quality on their business, or, at best, being reactive to data issues as they arose. A recent academic survey (Laranjeiro, Soydemir, Bernardino, 2015) investigating poor quality data highlighted the need for further research in the area of data quality.

The research discussed in this paper investigated how probability models might be used in a fundamental manner to evaluate the quality of data in a given dataset with no domain-specific criteria. The scope of the research is constrained to specific datasets and serves as an entry point into a larger body of research being undertaken into the field of data quality.

## 2 OUTLINE OF OBJECTIVES

The main objective of this research was to examine if probability models can be used to improve indicators of data quality in a given dataset, with no domain specific knowledge. Using supervised machine learning techniques, the hypothesis was that samples of known 'bad data quality' could be used to create models that could be subsequently applied to unseen data to provide an indicator if new data under examination represented good or bad quality data.

The specific objectives identified as key to guiding and performing the research effectively were the following:

- Design a methodology to investigate the research hypothesis.
- Research the state of the art and determine if there are any potential gaps in the literature in relation to the hypothesis.
- Select and explore a machine-learning technique to determine if one approach to investigating the hypothesis is more effective than others.
- Evaluate the effectiveness of the selected machine-learning techniques.
- Output the results of the research in the form of a research paper and submit in time for inclusion and publication in the Doctoral Consortium at the 9th International Conference on Data Science Technologies and Applications (DATA) and the 15th International conference on Software Technologies (ICSOFT), 2020.

## 3   STATE OF THE ART

The research to determine the state of the art was carried out using both open online resources, and open access resources retrieved through the university library. In addition, communication with fellow researchers and lecturer advisors assisted in identifying appropriate sources of information.

The current research is varied and distributed over a spectrum that covers different aspects of data quality which include general data or information quality (Loshin, 2010), investigations into data cleansing (Krishnan, et al 2016), the specific needs of data quality in relation to big data (Batini et al, 2016, Serhani et al, 2016, Abdullah et al, 2015), and specific approaches to more focused areas of data quality, for example entity resolution (Talburt, 2011). Other research focuses on model or framework approaches to the problem, for example Taleb et al (2016) who propose a Big Data evaluation scheme for Data Quality using sampling, and Li, Peng & Kennedy (2018) who define a rule based taxonomy of dirty data.

Apart from general analysis using data, other factors have focused a new lens on data quality, including the reliance of AI and Machine Learning on high quality input (Sessions & Valorta, 2006) and legislation (Hoeren 2018). Research relating to Data Quality in the field of Machine Learning normally focuses on aspects of missing data (Sessions & Valorta, 2006), but this is only one of the dimensions of Data Quality that have been identified by many (Batini & Scannapieco 2016) including the Data Management Association (DAMA, 2008). Data quality is a multi-dimensional concept, and organisations must deal with the objective measurements based on the data set in question (Pipino, Lee & Wang, 2002).

Woodall et al (2014) have examined the on-going trend of Big Data analytics and find that 'new Big Data paradigm opens new questions about the role of data quality management that need to be addressed'. Other researchers suggesting the need for further research in the area include Cai & Zhu (2015), Hashem et al (2015), Liu et l (2016) and Batini et al (2016).

One of the common themes identified throughout the literature is that in order to be able to determine what is good or bad quality data, we need to have some domain knowledge, or set of rules against which the data can be compared and evaluated. Much of the literature that refers to this describes 'data dimensions' as being the optimal way to view data for measurement. The most important dimensions of data quality relate to the accuracy, consistency, completeness, and correctness of the data under investigation (Fox, Levitin, Redman, 1994). A seminal paper on data quality (Pipino, Lee, Wang, 2002) describes fundamental principals to produce metrics against which quality of different data dimensions may be measured and data can be evaluated. The authors of this paper also propose three functional forms that they found are pervasive across different dimensions, including simple ratio, minimum or maximum operation, and weighted average.

This literature review has identified a gap in the research in relation to the overall approach taken by other researchers. The review suggests that for the papers examined, the literature always refers to one or more aspects of data dimensionality and/or domain knowledge. Little work has been identified that attempts to evaluate data quality without taking domain knowledge or data dimensionality into consideration. This provides justification for the experiment in this paper, which forms part of a larger body of work that will be the next stage of the authors main research.

## 4   METHODOLOGY & OUTCOME

The following methodology was utilised for the experiments:
- Establish experiment boundaries
- Choose dataset for testing
- Select machine-learning technique to use in the experiment and justify the choice
- Design an algorithm to produce sample 'bad quality data' for a given dataset to become a training dataset
- Write code to run the experiment by executing the chosen machine learning technique against the dataset
- Evaluate the results of the experiment
- Write up the results

### 4.1 Experiment boundaries

The experiment is restricted to pre-determined machine learning techniques and pre-selected datasets and is designed to both test the hypothesis and usefulness of the technique, excluding any reference to domain-specific criteria or data dimensionality.

## 4.2 Dataset choice

The dataset chosen was from the Kaggle data repository. There was no particular regard to the domain of the dataset chosen, the main criteria for choice was that there was a small number of fields in the schema to facilitate efficient processing of data from a model creation and analysis point of view. The chosen dataset was 'avocado prices 2016'. A sample of this data is shown in Figure 1.

| AveragePrice | Total Volume | Total Bags | Small Bags | Large Bags |
|---|---|---|---|---|
| 1.52 | 73341.73 | 11432.09 | 11017.32 | 411.83 |
| 1.53 | 68938.53 | 9504.66 | 8876.65 | 587.73 |
| 1.49 | 71777.85 | 12822.02 | 12176.75 | 645.27 |
| 1.48 | 113031.96 | 6704.29 | 6476.12 | 228.17 |
| 1.52 | 58171.89 | 8253.58 | 7973.98 | 279.6 |
| 1.56 | 70089.51 | 9504.34 | 9238.4 | 209 |
| 1.62 | 63608.01 | 10212.7 | 9790.67 | 422.03 |
| 1.63 | 57178.2 | 10905.44 | 10474.09 | 431.35 |

Figure 1: Original data sample.

## 4.3 Machine learning technique selection

This experiment was designed to identify a binary difference between the data under examination (this being good vs bad quality data), however, it was expected that there would be some overlap between results due to unknown variability resulting from domain specific data. For this reason, it was decided to choose a more sophisticated approach for this initial experiment.

The method chosen was decision trees, using the Catboost library (Dorogush, Ershov, Gulin, 2018). Decision trees are a clustering method and chosen based on the hypothesis that good and bad data points would share similar attributes, and therefore should tend to cluster together. In addition to standard clustering one would expect from for example K-Means, a decision tree is capable of handling data with high rates of missing data and errors. The Catboost library supports categorical and numerical features, and its high-performance gradient boosting has proven extremely capable in comparison to alternatives.

## 4.3 Algorithm design

The chosen dataset from the Kaggle repository was a collection of prices of avocados during 2016. In order to introduce corrupted data, an algorithm was designed to iterate through the data in each column and extract the minimum and maximum values. Once this was done, the dataset was again iterated and values in selected indexes changed to include different values, emulating outlier data. These changed rows of data were then labelled as 'bad' with unchanged rows marked 'good'. Figure 2 illustrates sample data after randomised labelling.

| AveragePrice | Total Volume | Total Bags | Small Bags | Large Bags | label |
|---|---|---|---|---|---|
| 1.52 | 73341.73 | -11425.07 | 9978555.66 | 411.83 | bad |
| 1.53 | 68938.53 | -9497.64 | -8876.65 | 3373665.6 | bad |
| 1.49 | 71777.85 | 12822.02 | 12176.75 | -645.27 | bad |
| 1.48 | 113031.96 | 6704.29 | 6476.12 | 228.17 | good |
| -1.01 | 58171.89 | 8253.58 | 9975512.32 | 279.6 | bad |
| 1.56 | 70089.51 | 9504.34 | 9238.4 | 209 | good |
| 1.62 | 63608.01 | 10212.7 | 9790.67 | 422.03 | good |

Figure 2: Changed data sample.

The following is the code used to implement the algorithm described.

```
import pandas as pd
import random

df = pd.read_csv('prices_original.csv')

max_values = {c:df[c].max() for c in
df.columns}
min_values = {c: df[c].min() for c in
df.columns}

df['label'] = 'good'
df_columns = [c for c in df.columns if
c != 'label']
for _ in range(len(df) // 2):
    index =
random.choice(range(len(df)))
    offset = random.randint(2, 10)
    column = random.choice(df_columns)
    df[column].at[index] =
max_values[column] +
df[column].at[index]
    df['label'].at[index] = 'bad'

for i in range(len(df) // 2):
    index =
random.choice(range(len(df)))
    offset = random.randint(2, 10)
    column = random.choice(df_columns)
    df[column].at[index] =
min_values[column] -
df[column].at[index]
    df['label'].at[index] = 'bad'

df.to_csv('prices_corrupted.csv',
index=False)
```

## 4.4 Experiment logic

Modern machine learning libraries have significantly reduced the amount of work required to setup and preform an experiment, and the Catboost library is no exception in this regard. To set up the experiment, the corrupted dataset was taken as the primary input. The input data was then split into two parts, representing known good, and bad quality data. This could be achieved easily as the previous data corruption process had changed data randomly and differentiated good from bad quality data using labels.

The model was established by creating an instance of a CatBoostClassifier and configuring it to use a specific number of iterations and instruct it to use either the CPU or GPU for processing. Training of the model was carried out by passing each data point into the model-fit method and flagging the training according to label associated with the data point. With the model having been trained, predictions could then be made on remaining data into a new column 'predictions', into which a label of 'good' or 'bad' was placed. Figure 3 illustrates the output generated once the experiment has been executed in a grid. Figure 4 illustrates the result in a chart.

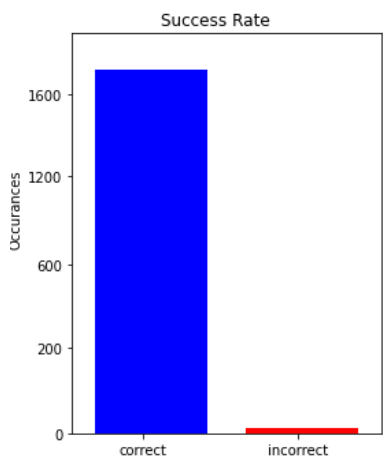

Figure 3: Predicted output data as grid sample.

Figure 4: Predicted output data as chart sample.

The following is the code used to implement the experiment logic described.

```
'''
Gradient boosting classification using
the Catboost library.

To install requirements:
pip install catboost
pip install pandas
'''
import pandas as pd
from catboost import CatBoostClassifier

OUTPUT_FILENAME =
'dataquality_predictions.csv'
DATA_FILENAME = 'prices_corrupted.csv'
TRAINING_ITERATIONS = 500

if __name__ == '__main__':
    # Load datasets.
    df = pd.read_csv(DATA_FILENAME)

    # Naive train/test split. This
works because we know that we corrupted
our
    # data randomly so there is no
difference which part of the dataset we
are going
    # to use for training and testing.
    train = df.iloc[:3300]
    test = df.iloc[3300:]
    # Initialize and train the model.
    model =
CatBoostClassifier(iterations=TRAINING_
ITERATIONS, task_type='CPU')
    model.fit(train[[c for c in
train.columns if c != 'label']],
train['label'])
    # Generate predictions.
    predictions = model.predict(test[[c
for c in test.columns if c !=
'label']])
    # As we sliced the dataset for
training and testing
    # we should now copy it in order to
add the predictions column.
    test = test.copy()
    test['prediction'] = predictions
    test.to_csv(OUTPUT_FILENAME,
index=False)
```

### 4.5 Results evaluation

The results of the research using the probability model provided by the Catboost library in this very limited experiment had a successful prediction of good versus bad quality of 97%. While this may seem like a good result, in real world situations with data, context is usually extremely important. The algorithm used to create pseudo corrupted data was naïve, and did not take either domain-specific indicators or data dimensionality into account. In addition, this is also only one result against a single dataset using a single machine learning approach. While the ability to predict good or bad data based on pre-labelled learning examples is accurate, in general it may not be sufficient for production-quality data that is rich with domain-specific metrics.

## 5 CONCLUSIONS

The research available that focuses on domain specific requirements in relation to evaluation of data quality is significant. Work carried out in relation to the dimensionality of data is also very prevalent. The term Data Quality is often described as "fitness for use" (Wang and Strong 1996) but can mean different things in different domains. Research in specific areas has been carried out in numerous domains (Gudivada et al 2017), and a range of domain specific techniques have been developed to evaluate and improve Data Quality (Batini et al 2009, Ganti & Sarma 2013, Loshin 2010).

It is acknowledged that while naive approaches such as the one described in this paper may work in certain cases, it is far from clear how well such methods would work in production environments across multiple domains. Further work is clearly needed to be able to properly evaluate the methods and techniques described in the experiment and this will be the focus of future research.

## 6 STAGE OF THE RESEARCH

This paper has outlined the start of a single experiment that forms part of a larger body of research being undertaken as part of a PhD path. The ongoing research will further investigate the hypothesis described in this paper, and also seek to investigate how machine learning techniques can be applied to improve data quality while also considering the data dimensionality and domain specific data requirements.